\documentclass[10pt,twocolumn,letterpaper]{article}

\usepackage{wacv}
\usepackage{times}
\usepackage{epsfig}
\usepackage{graphicx}
\usepackage{amsmath}
\usepackage{amssymb}
\usepackage{footnote}
\makesavenoteenv{tabular}
\makesavenoteenv{table}
\usepackage{balance}

\wacvfinalcopy 

\def\wacvPaperID{331} 

\ifwacvfinal\fi
\begin{document}

\title{Confidence Prediction for Lexicon-Free OCR}

\author{Noam Mor \hspace{2cm} Lior Wolf \\
The School of Computer Science\\
Tel Aviv University\\
{\tt\small noam.mor@cs.tau.ac.il}
}

\maketitle
\ifwacvfinal\thispagestyle{empty}\fi

\begin{abstract}
Having a reliable accuracy score is crucial for real world applications of OCR, since such systems are judged by the number of false readings. Lexicon-based OCR systems, which deal with what is essentially a multi-class classification problem, often employ methods explicitly taking into account the lexicon, in order to improve accuracy. However, in lexicon-free scenarios, filtering errors requires an explicit confidence calculation. In this work we show two explicit confidence measurement techniques, and show that they are able to achieve a significant reduction in misreads on both standard benchmarks and a proprietary dataset.
\end{abstract}

\section{Introduction}

Optical Character Recognition (OCR) is the task reading a natural image of printed or hand-written text. This task has many applications, such as automatic processing of documents and forms, automatically routing envelopes based on zip code, and reading aloud photographed text for the visually impaired. Additionally, automatic text recognition is a necessary step in many other tasks, such as reading the speed limit as part of the automatic driving task.

Recent years have shown major advancement in OCR accuracy. Literature pertaining to providing a confidence measure for these systems has, however, been relatively sparse. Real-world usage of OCR systems requires such a confidence measure, since the last stage in an industrial OCR pipeline is typically paying a human typist to type out the content of a given image that the system rejected. As such, real-world OCR systems are measured by how much of their input can be processed without human intervention, while keeping the error level below a certain threshold. In such a setting, assuming that the acceptable error level is 1\%, a system that achieves an overall accuracy of 90\% but contains a mechanism that can guarantee an error level of 1\% on its most confident 50\% of the input, is more valuable than a system that provides 98\% accuracy but provides no confidence score at all.

Most OCR systems in the literature are limited to a predefined lexicon, in either the training phase, the inference phase, or both. A predefined lexicon may be very helpful for systems that are concerned with recognizing natural language text, as typically all words in the dataset will be derived from a limited dictionary. An example of a lexicon-based method is using a language model to only produce outputs that are likely words in the target language. Another example is mapping the image space and the label space to a common  feature space, and outputting the nearest dictionary label for a given input image.

Lexicon-based OCR methods often contain intrinsic confidence measures - a language model defines the probability that a label is a legal word in the language, and the distance between the input and the output in a shared feature space may be interpreted as the confidence that the output matches the input.

Some datasets, however, are incompatible with the idea of a limited lexicon. Such datasets include scanned fields containing social security numbers, photographed license plates, any kind of texts whose origin is unknown, and others. For such a dataset, an explicit language model cannot be constructed, since the ability to predict the next character given the previous is very limited. Likewise the usefulness of a distance measurement between an image and a label in a learned shared space is limited, since the nearest neighbor search would be performed between what is essentially all possible labels, which is impractical. When dealing with such datasets, for the purpose of providing a confidence measure of the OCR labeling, a lexicon-free confidence measure must be devised.

This work is concerned with providing a confidence score to an OCR system whose architecture is inspired by state-of-the-art works in the OCR field. In particular, we base our network on Liu \etal~\cite{liu2016star}, which achieves state-of-the-art results in several accuracy benchmarks, but provides no confidence measurement outputs. 

Our main contribution is the presentation of two techniques for obtaining a confidence measure, that do not require any information about the lexicon, and so are also applicable in a lexicon-free scenario.

The first of the two techniques is inspired by a common technique in multi-class classification. We obtain the probabilities that the OCR model assigns to its first and second most probable readings, given an input image. We then give a high confidence score where the first reading is much more probable than the second reading. If it is not, then the sample is ambiguous, and so its confidence score is lowered. Our experiments show that this confidence measurement is highly effective at detecting errors in the OCR model.

While the first technique proved to be effective as an error predictor, real-world concerns dictate a more involved solution. A word- or a phrase-level OCR system is typically one component of a pipeline, that often includes alignment, segmentation, et cetera. Since previous components are not always accurate, a sample may be passed to the OCR system, that the system must reject. Therefore, a requirement of an OCR system in a real-world scenario is to be able not only to provide accurate output, but also to be able to reject faulty samples.

Our second error prediction mechanism is designed to handle this use-case. We add an error prediction branch to the OCR network, which directly predicts  whether either the OCR model will err given the input, or the sample should be rejected. Our experiments show that when sufficient training data is available, this learned prediction model is able to detect OCR errors with a high degree of accuracy, as well as detect samples that must be rejected. We show that on a proprietary dataset by the document processing automation company Orbograph, which includes such reject samples, our second method outperforms the first.

Another advantage of our error-prediction learning method is that it does not depend on the particular decoding component of the network, which makes it applicable to the vast majority of the OCR neural network architectures.

This work is structured as follows. In the following section, we overview existing literature in the field of OCR and OCR confidence scoring. Section~\ref{section:methodology} presents our system in detail. Section~\ref{section:experiments} details the experiments we have conducted and the datasets we have used. We conclude in Section~\ref{section:conclusions}.

\section{Prior Work} \label{section:prior}

Where applicable, lexicon-based OCR is able to beat lexicon-free OCR in accuracy by a large margin. A few select examples of neural-network-based methods incorporating a lexicon into their OCR networks follow. 

Shi \etal~\cite{shi2017end} introduce a two-stage lexicon-based decoding during testing. In the first stage, a lexicon-free decoding method is employed to obtain a candidate output $l$. Then, their system restricts itself only to lexicon items within a certain edit distance $\delta$ of $l$, and finds the most likely one as its final output. 

Ghosh and Valveny~\cite{ghosh2017visual} also incorporate the lexicon into their system's decoding stage. They train an explicit n-gram character-level language model on the lexicon, and during beam-search decoding, the OCR word probability is weighted with the language model's word probability. This technique greatly increases the probability that the model outputs a word from the lexicon, and enables the decoding phase to ignore possible readings that are not legal words in the language.

Poznanski and Wolf~\cite{poznanski2016cnn} explicitly treat the word OCR task as a multi-class classification problem. During training, their system learns to predict from an image a vector of binary textual features computed from the image's label. A regularized Canonical Correlation Analysis model is then trained between the network's neural code and the label's textual feature vectors, to bring both vectors to a shared space and neutralize cross-correlations between different textual features. At test time, the final output is determined by a nearest-neighbor search of an image's neural representation, to obtain the lexicon entry closest to the image in the common learned space.

Most OCR systems in the literature are capable of being used without a lexicon - in particular, those based on the CTC loss, which we use, and will be described in detail in Section~\ref{section:ctc}. Several works on lexicon-based OCR also publish lexicon-free accuracy results, as in \cite{shi2017end,wang2011end,liu2016star,he2016deep}.

As of writing these lines, the current state of the art in several OCR accuracy benchmarks, both constrained and lexicon-free, is the STAR-net network by Liu \etal~\cite{liu2016star}. Their system is comprised of a convolutional image-encoding stage, followed by a deep LSTM network, and trained with the CTC loss. They also employ residual connections within their convolutional model, and a spatial transformation stage to rotate scene-text to a canonical perspective. 

The literature on OCR confidence is relatively sparse. However, virtually all commercial OCR engines do provide a confidence score. We examine the well-known and open-source Tesseract \cite{smith2007overview} OCR engine. While Tesseract does have some limited lexical analysis, it does not use its lexicon information for calculating its confidence score. 

We now give a brief overview of the method Tesseract uses for its confidence calculation. During its OCR phase, text is segmented into words, and words are segmented into characters. Each detected character is then matched to a character \emph{prototype}, which is a canonical shape of a character. This matching is done by extracting a visual feature vector from the detected character's contour, and searching for the closest character prototype in feature space. The shortest such distance is then called the confidence for the character. These character confidence scores are then weighed and summed to obtain a word rating, which serves as the overall confidence score for the word.

\begin{figure*}[t]
  \includegraphics[width=.95\linewidth]{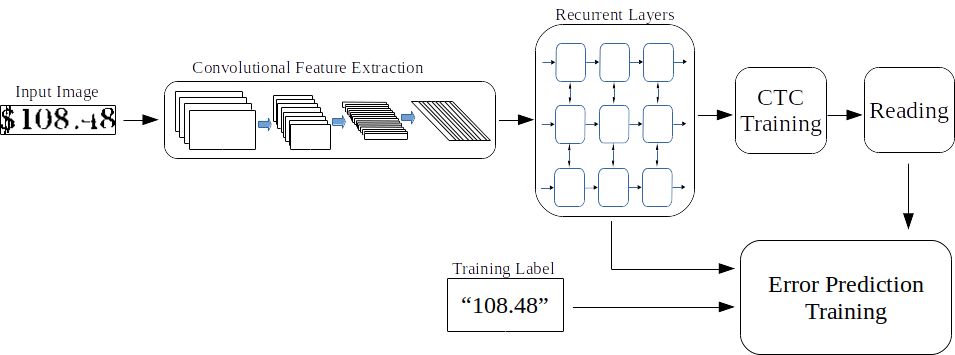}
 \smallskip
\caption{A schematic diagram of our system. For training, pairs of images and labels must be provided. The image goes through a convolutional stage, followed by three-layer deep bidirectional LSTM. The output from the recurrent layers is input to the CTC loss. Additionally, the most probable reading from the OCR model is compared with the training label. The output from an intermediate LSTM layer serves as input to an error prediction model, which predicts whether the most probable OCR reading matches the training label.}
  \label{fig:net_diagram}
\end{figure*}

Another method used by commercial OCR systems is trying to assess the image quality, and giving lower scores for low image quality.

\section{Methodology} \label{section:methodology}

This work focuses on the task of text recognition in a cropped, single-line
image. In the following sections, we describe the main components of our system: the text recognition model and the dedicated error prediction component.

\subsection{OCR Model}

Our text recognition model includes convolutional visual feature extraction,
followed by recurrent layers, and a final transcription layer.

\subsubsection{Visual Feature Extraction}

For visual feature extraction, following~\cite{liu2016star}, we employ
a convolutional neural network with residual connections~\cite{he2016deep}.
The input to the network is an image tensor of dimensions $W\times64\times1$,
where $W$ is a varying image width. The network is comprised of 6
blocks of 2 residual convolution modules, where each residual module
is composed of two convolutional layers, a residual connection and
batch normalization. The convolutions use a kernel of size $3\times3$
and padding $1$.

Interspersed between the blocks are max pooling operations, with stride
of either $2\times2$ or $1\times2$, which serve to reduce the spatial
dimension of the image. We increase the number of filters in later
blocks by a factor of 2. After the visual feature extraction, we get
an activation tensor of dimensions $T\times1\times512$, where $T=\frac{W}{8}$.
We treat this tensor as $T$ visual feature vectors of length $512$. 

\subsubsection{Recurrent Layers}

Each visual feature vector is the result of applying a sequence and
convolutions and pooling operations to part of the input image. As
such, the feature vectors describe the image in a sliding window,
and therefore are spatially local. To introduce time-dependence between
the feature vectors, we apply 3 layers of bi-directional LSTM on the
feature vector sequence, with $512$ units in each direction. 

The output of the final LSTM is then projected to $\left|\Sigma\right|+1$
dimensions, where $\Sigma$ is the set of possible characters in the
output text (the alphabet). These serve as character probability distribution
for the transcription layer.

\subsubsection{Transcription Layer} \label{section:ctc}

We adopt the Connectionist
Temporal Classification (CTC) layer proposed by~\cite{graves2006connectionist}.
We describe it briefly. 

Let $\Sigma$ the alphabet, and $\Sigma'=\Sigma\cup\left\{ -\right\} $, where $-$ is a special
``blank'' character. CTC defines a transcription function $\mathcal{B}:\Sigma'^{*}\rightarrow\Sigma^{*}$
as follows: first, replace each repeated character by a single character;
then remove all blank characters. For example, $\mathcal{B}$ maps
the string ``hhee-{}-ll-lo-{}-'' to ``hello''.

CTC then defines the probability for the string $l$ given a sequence
of probability distributions $y=y_{1}\ldots y_{T}$ as the probability
that $\mathcal{B}\left(a_{1}\ldots a_{T}\right)=l$, where $a_{1}\ldots a_{T}$
are independently sampled from from $y_{1}\ldots y_{T}$, respectively.
Formally:
\begin{equation}
p\left(\pi|y\right)=\prod_{t=1}^{T}y_{\pi_{t}}
\end{equation}
and:
\begin{equation}
p\left(l|y\right)=\sum_{\pi:\mathcal{B}\left(\pi\right)=l}p\left(\pi|y\right).
\end{equation}
Naive computation of this sum is infeasible, since the number of elements
in the sum grows exponentially with $T$. Fortunately, this probability
and its gradient can be computed efficiently by the forward-backward
algorithm defined in~\cite{graves2006connectionist}. Since the CTC
conditional probability is differentiable, it can be directly optimized
using gradient descent. The optimization objective is then defined
to be the negative log-likelihood:
\begin{equation}
\min\qquad-\sum_{i}\log p\left(l_{i}|y_{i}\right)
\end{equation}

Since the CTC conditional probability depends on the probability vectors and the ground truth label, the image labels need only include the image text, and thus any positional information is unnecessary. This being the case, OCR systems based on this model are structurally much simpler than traditional OCR systems, as there is no intermediate segmentation stage.

At test time, a beam prefix search on $y$ is performed to find the $k$ most probable labels given $y$, and their respective probabilities.

\subsection{Probability Ratio As Confidence}

\begin{figure}
\center
  \includegraphics[width=0.45\textwidth]{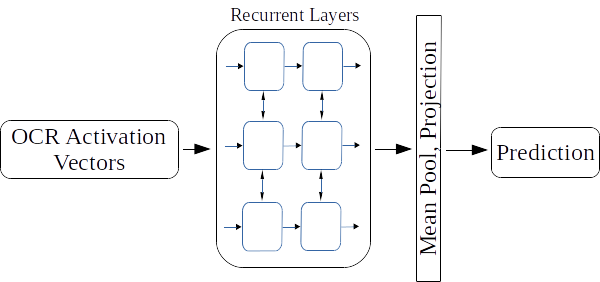}
 \smallskip
\caption{A schematic diagram of the error prediction sub-model. Activations from an intermediate stage inside the OCR model are input to multi-layer deep bidirectional LSTM. The resulting vectors are mean-pooled and projected to obtain the log-probability for the error prediction. This sub-model is trained with the binary classification loss, with the label being whether the OCR model was right on current sample.}
  \label{fig:net_diagram}
\end{figure}

Providing a confidence measure for the text recognition model is a crucial part of this work. One might naturally expect the conditional probability that the CTC calculation assigns to the label by itself would be able to serve as a good measure of confidence. However, testing has found that it performs poorly as such.

Our model, however, is capable of outputting more than one possible label per input image, and assigning probabilities for all of them. In particular, we utilize the probabilities assigned to the first two most probable readings.

Our first error prediction method is as follows. Let $x$ be the input image, $l_1$ and $l_2$ the two most likely readings from the CTC model, and $p_1$ and $p_2$ the conditional probabilities $\Pr\left[l_{1}|x\right]$ and $\Pr\left[l_{2}|x\right]$. Then the confidence score we assign to $l_1$ is:

\begin{equation}
\textit{confidence}=1-\frac{p_{2}}{p_{1}}.
\end{equation}

By definition, we have that $p_1 \ge p_2$, and so this expression gives a number in the range $\left[0, 1\right]$, as is desirable for such a score. When $p_1 \gg p_2$, our system is the most confident, and the $\mathit{confidence}$ value will be close to 1. 

This method is both easy to implement, and gives excellent results in detecting errors. We show detailed benchmarks results in Section~\ref{section:experiments}.

\subsection{Error Prediction as Confidence} \label{subsec:Directly-Predicting-Rejects}

Whereas the previously described method is effective at error prevention, it is inadequate for recognizing faulty input images that need be rejected. This is to be expected, as the set of common invalid image inputs is highly dependent on the specifics of a system, as well as the dataset. We believe that industry-grade OCR systems need to be specifically trained to recognize their own faulty images.

It is generally desirable to have a single accept-or-reject confidence score. Therefore our second method is a single confidence mechanism, that learns to predict both whether the OCR branch will give the correct labeling, and whether a given sample should be rejected.

Another advantage of this method is that it does not assume that the corresponding OCR network is trained using CTC. It should be straightforward to re-use the error prediction formulation defined ahead in an OCR system that has a different decoding phase.

Our approach is the following: we postulate that the confidence measure for a prediction $\hat{l}$ given image $x$ is the probability $p$ that $\hat{l}=l$ and $x$ should not be rejected, where $l$ is the real label. We then proceed to train a binary classification model that given $x$, predicts the probability for this event.

The error prediction branch takes as input the activations from an intermediate stage in the network, and its output is the probability that the OCR output is incorrect. Specifically, we take the activations of the first bi-LSTM layer and apply, in this new branch, two 512-unit bidirectional LSTM layers. The LSTM's output vectors are then projected to dimension 1 to obtain a vector $v\in\mathbb{R}^{T}$. The final probability is $\sigma\left(\frac{\Sigma v}{T}\right)$, where $\sigma$ is the logistic function. Division by $T$ is performed to neutralize the impact of the image width on the prediction result.

This branch is trained by supplying an image $x$, a label $l$, and the reject label $r$. The OCR model is run on $x$ to obtain a predicted label $\hat{l}$. We calculate the desired output of the error prediction as:

\begin{equation}
y=\mathcal{I}\left[l \ne \hat{l} \vee r \right].
\end{equation}
The error prediction branch is then trained to minimize the cross-entropy loss:
\begin{equation}
\min\qquad-\sum_{i}\left[y_{i}\log p_{i}+\left(1-y_{i}\right)\log\left(1-p_{i}\right)\right].
\end{equation}
A curious problem arises whenever the training dataset is small enough to reach overfitting. In such a situation, it is necessary to dedicate a subset of the training data to the training of the error prediction branch exclusively. The reason is that when the OCR network overfits, it outputs the correct label on all training samples, and the error
prediction branch learns that the OCR network is never wrong. If the dataset is big enough to avoid overfitting, then both branches can be trained simultaneously.

\section{Experiments} \label{section:experiments}

To evaluate the effectiveness of the proposed error prediction methods, we conducted experiments on standard benchmarks for scene text recognition and for handwritten text recognition. Additionally, we conduct experiments on a proprietary dataset provided to us by Orbograph, of real-world cropped images of scanned dollar amounts.

For each dataset, we present, in addition to our lexicon-free accuracy, a ROC curve, which shows the trade-off between the misread rate and the total system read rate for each method. For example, a point at 0.2 misread rate and 0.9 read rate denotes that by picking a certain confidence threshold, the system rejects 10\% of the valid data, and reduces the number of errors by 80\%. To compare the effectiveness of different confidence methods, we use the area-under-curve (AUC) metric.

We compare the results of both our methods with two baselines. The first is using the CTC probability $p$ itself as a confidence measurement. As mentioned above, $p$ tends to decrease as the image gets wider. Our second baseline, $p^{-T}$, is an attempt at normalizing this relation in a simple manner, and trying to get a better baseline than just $p$.

On large datasets, we trained the error-prediction sub-network simultaneously with the OCR sub-network, and used the same data as training data. On the handwriting datasets, which are substantially smaller, we dedicated part of the training set to training only the error predicting sub-network. 

\subsection{Scene Text Datasets}

For all scene text recognition experiments, we follow common practice of using the large synthetic dataset by Jaderberg \etal~\cite{jaderberg2014synthetic},
whose training set contains 8 million synthetic training samples. We train a single network on the synthetic dataset, and use that network for all scene text recognition benchmarks, without any fine-tuning.

We evaluate our scene text performance on two popular benchmarks - ICDAR 2013 (IC13), and Street View Text (SVT).
\begin{description}
\item [{IC13~\cite{karatzas2013icdar}}] the test set contains 251 scene images with labeled bounding boxes. Following Wang \etal~\cite{wang2011end}, we restrict the test set to contain only alpha-numeric characters with three characters or more, resulting in a test set of about 900 cropped text images.
\item [{SVT~\cite{wang2011end}}] consists of 249 images collected from
Google Street View. The word images have been cropped from these images,
resulting in a test set of 647 samples.
\end{description}

\subsection{Handwritten Datasets}

We conduct experiments on two well-known dataset for handwriting recognition - IAM and RIMES. Since these datasets are very small, we used the validation set to train the error prediction branch.
\begin{description}
\item [{IAM~\cite{marti2002iam}}] dataset consists of scanned handwritten
English texts by 657 writers, giving a total of 115,320 labeled word images. After filtering out punctuation and short words, we obtain
a training set of 40,526 samples.
\item [{RIMES~\cite{grosicki2009icdar}}] dataset consists of scanned handwritten
French texts by 1300 people, giving a total of 67,000 words. Restricted to long enough words without punctuation, the training set size becomes 35,723 samples.
\end{description}

\subsection{Orbograph's Dollar Amount Dataset}

In addition to the above standard benchmarks, we evaluated our method on a proprietary dataset by Orbograph. This dataset consists of 1 million labeled images of automatically scanned and cropped fields denoting dollar amounts. This dataset was created by collecting dollar amount scanned text samples from real-world printed documents and having them manually labeled. Since this is unfiltered real-world data, it has an amount of reject samples, which we use to train our second error prediction method, described in~\ref{subsec:Directly-Predicting-Rejects}.

In Figure~\ref{fig:res2}, we provide a ROC curve on the test set. A misread refers to either a reject image receiving a high confidence score, or a non-reject image receiving the wrong transcription with a high confidence score. We provide an additional ROC curve for the non-reject samples only. 

\subsection{Technical Details}
Our system was implemented in TensorFlow and trained on an nVidia Titan X and on an nVidia K80. CTC decoding was used a search beam of size 100. Training was done using ADAM\cite{kingma2014adam}, with an initial learning rate of $10^{-4}$.

\subsection{Results}

\begin{table*}[t]
\centering
\begin{tabular}{|c|c|c|c|c|}
\hline 
 & SVT & ICD13 & IAM & RIMES\tabularnewline
\hline 
\hline 
Accuracy & 78.67\% & 88.91\% & 79.51\% & 88.05\%\tabularnewline
\hline 
CTC AUC & 0.484 & 0.463 & 0.493 & 0.544\tabularnewline
\hline
CTC (norm) AUC & 0.516 & 0.445 & 0.461 & 0.555\tabularnewline
\hline
CTC ratio AUC (ours) & \textbf{0.937} & \textbf{0.965} & \textbf{0.913} & \textbf{0.949}\tabularnewline
\hline 
ErrPredict AUC (ours) & 0.891 & 0.941 & 0.793 & 0.818\tabularnewline
\hline 
\end{tabular}
\smallskip
\caption{The accuracy obtained for each literature benchmark, as well the the AUC obtained by the baselines and the proposed confidence methods.}
\label{tab:results1}
\end{table*}
\begin{table*}
\centering
\begin{tabular}{|c|c|c|}
\hline 
 & Orbograph (all) & Orbograph (No rejects)\tabularnewline
\hline 
\hline 
Accuracy & 96.75\% (*) & 99.75\%\tabularnewline
\hline 
CTC AUC & 0.537 & 0.529\tabularnewline
\hline 
CTC (norm) AUC & 0.681 & 0.603 \tabularnewline
\hline
CTC ratio AUC (ours) & 0.987 & \textbf{0.987} \tabularnewline
\hline
ErrPredict AUC (ours) & \textbf{0.998} & 0.978\tabularnewline
\hline 
\end{tabular}
\smallskip
\caption{The accuracy obtained for the numeric dollar amount dataset as well the the AUC obtained by the baseline and the proposed confidence methods. (*) The accuracy value for Orbograph (all) refers to all reject samples as misses, and as such it serves to demonstrate the need of an accurate confidence signal.}  
\label{tab:results2}
\end{table*}

\begin{figure*}[t]
\begin{small}
\centering
  \begin{tabular}{cc}
  \includegraphics[width=.45\linewidth]{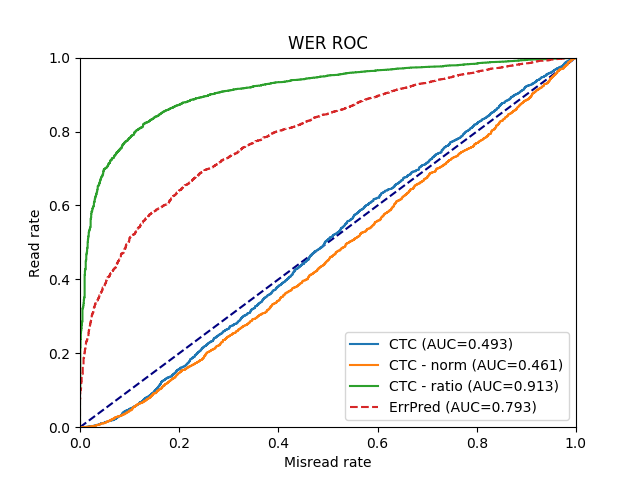}&
\includegraphics[width=.45\linewidth]{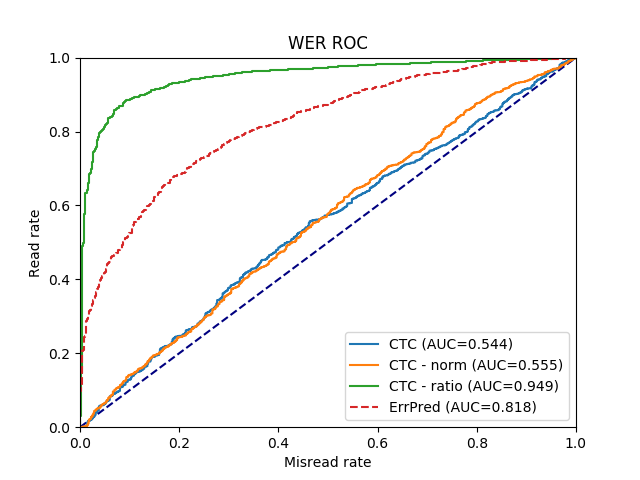}\\
 (IAM) & (RIMES)\\
 \includegraphics[width=.45\linewidth]{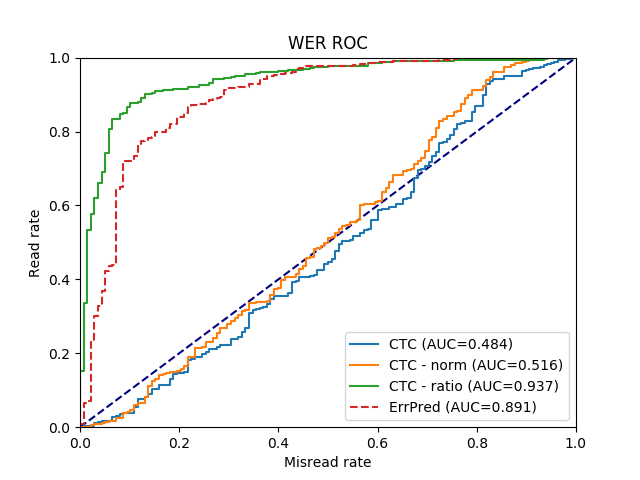}&
\includegraphics[width=.45\linewidth]{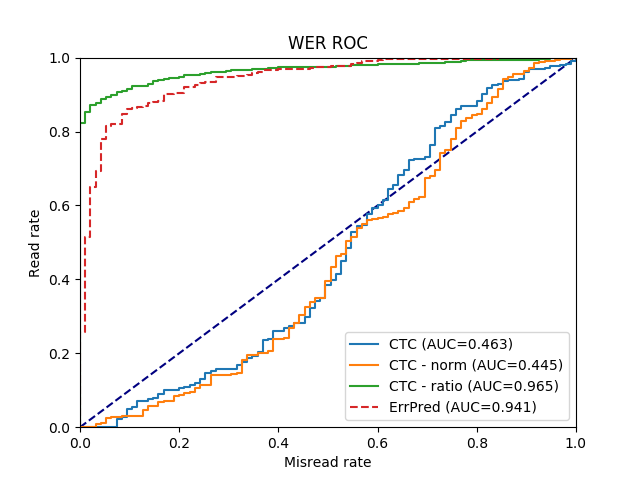}\\
 (SVT) & (IC	13)\\
\end{tabular}
 \end{small}
 \smallskip
\caption{ROC curves for the literature benchmarks for both the baseline CTC-based confidence score as well as our proposed methods.}
  \label{fig:res1}
\end{figure*}

\begin{figure*}[t]
\begin{small}
\centering
  \begin{tabular}{cc}
  \includegraphics[width=.45\linewidth]{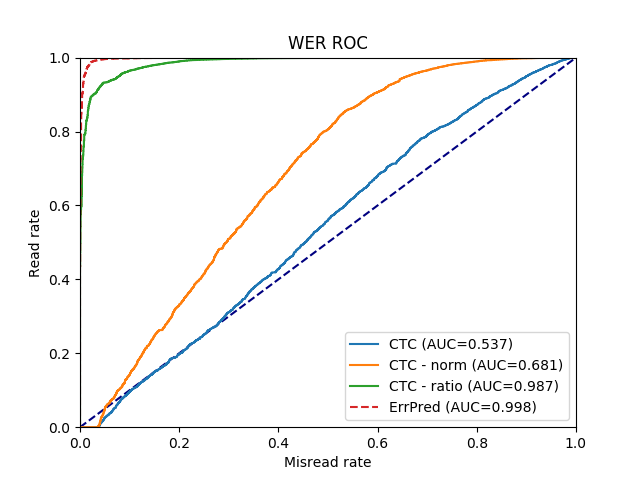}&
\includegraphics[width=.45\linewidth]{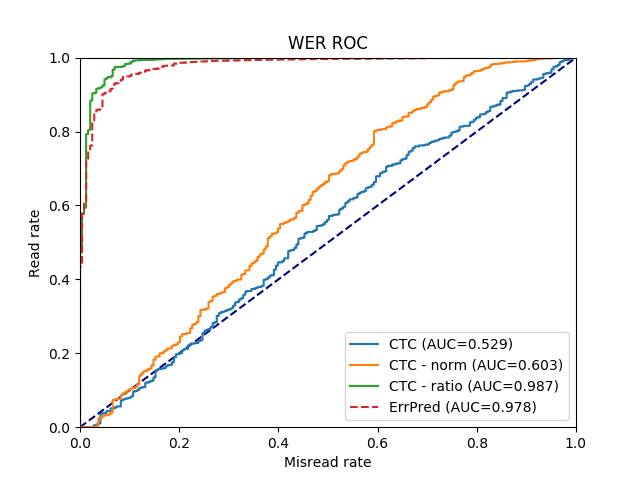}\\
 (Orbograph-All) & (Orbograph-No rejects)\\
\end{tabular}
 \end{small}
 \smallskip
\caption{ROC curves for the real-world numeric dollar amounts dataset for both the baseline CTC-based confidence scores as well as the our proposed methods.}
  \label{fig:res2}
\end{figure*}

\begin{figure*}[t]
\begin{small}
\centering
  \includegraphics[width=\linewidth]{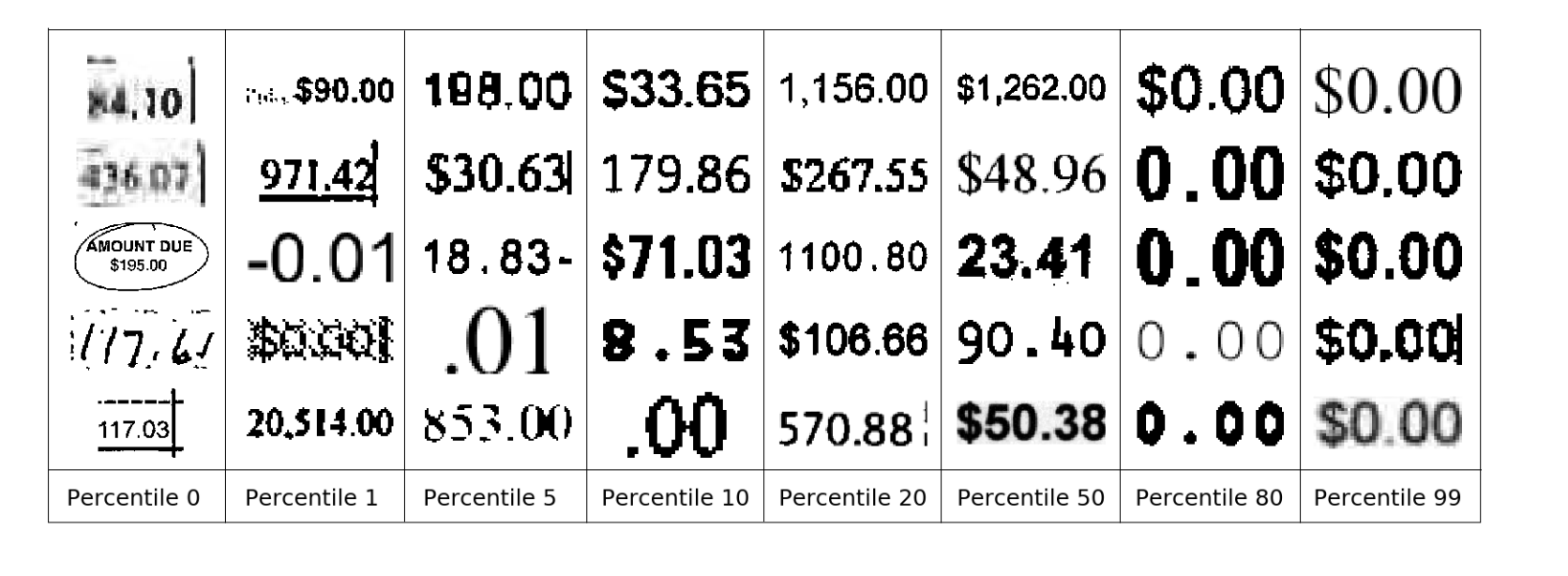}
 \end{small}
 \smallskip
\caption{Samples from Orbograph's dollar amounts dataset, grouped by percentile of confidence score given by the error prediction model. For example, the five samples under "Percentile 0" got a confidence scores in the lowest 1\%. One can see that the model correctly gives a low confidence score to difficult samples, such as unclear text, images with extraneous information, and handwritten text. Additionally, one can see that a large portion of the dataset has the label "0.00", and so the higher confidence percentiles are dominated by "0.00" samples.}
  \label{fig:qualitative}
\end{figure*}
 
Our results, depicted in Tab.~\ref{tab:results1} and~\ref{tab:results2} as well as in Fig.~\ref{fig:res1} and~\ref{fig:res2}, clearly show that both our methods are effective in predicting and preventing OCR errors. In all datasets that have no reject samples, our probability ratio method proves to be highly effective. As expected, on small datasets our learning-based error prediction method proved to be significantly worse than the probability ratio method, due to a lack of data. When training on large datasets, both methods obtain very convincing results.

Orbograph's dataset provides ideal benchmark conditions, as it is not synthetic and contains 1 million samples. Additionally, this dataset
exactly mimics the real-world conditions and requirements of an OCR system by containing reject samples. As shown in Fig.~\ref{fig:res2}, when measuring error prediction accuracy without considering rejects, we see that both methods give very good performance, with the probability ratio method getting a slight edge. However, when considering reject samples as well, our learning prediction model achieves exceptionally good results. The error prediction model is able to reduce the error rate by as much as 99.7\%, while rejecting as few as 1\% of the valid samples.

Figure~\ref{fig:qualitative} shows non-reject examples from Orbograph's dollar amount dataset at different levels of confidence, for qualitative assessment. We expect to find that samples from the bottom confidence percentile are difficult to read for an OCR model trained on cropped, printed dollar amounts. And indeed, one sample is handwritten, one includes the surrounding box due to imperfect cropping, one includes unrelated text and markings, and the other two are blurry. Text in the fifth percentile is already fully legible and reasonably cropped. 

It is also interesting to note that the top percentiles are completely dominated by samples with the label "\$0.00". The reason for that is that the dataset includes very many such samples, and so the model learned to recognize them with high confidence.

To conclude, our methods vastly outperform the baselines suggested, and prove to be applicable to printed text OCR, handwritten text OCR, and scene text recognition.

\section{Conclusions} \label{section:conclusions}

The best lexicon-free deep-learning OCR systems in the literature today use CTC for training. However, they contain no error detection mechanisms. As such, these methods cannot be adopted as-is in real-world systems. It is our observation that a lack of error-detection research has hindered the widespread adoption of modern deep-learning based OCR methods by the industry.

As our experiments show, a system can be built that is able to predict OCR errors directly from the probabilities predicted by CTC, with excellent accuracy. 

Additionally we propose a novel solution that modifies an OCR neural network by attaching a secondary deep network that is specifically aimed at learning a confidence measure for the main OCR network. Our error-prediction sub-network does not assume that the OCR branch is trained using CTC, and so it can be applied to virtually any OCR network that employs a CNN followed by recurrent layers.

In an extensive set of experiments we demonstrate that our methods are capable of achieving accurate error predictions. Our learned model achieves particularly noteworthy results on Orbograph's proprietary numeric dataset, where our system is able to achieve an accuracy two orders of magnitude greater than human typists, while keeping the read-rate at a very high level.

We believe our methods to be both simple to implement, and necessary for industry adoption of modern, deep neural-network based OCR methods. 

\balance
{\small
\bibliographystyle{ieee}
\bibliography{citations}

\begin{thebibliography}{10}\itemsep=-1pt

\bibitem{ghosh2017visual}
S.~K. Ghosh, E.~Valveny, and A.~D. Bagdanov.
\newblock Visual attention models for scene text recognition.
\newblock {\em arXiv preprint arXiv:1706.01487}, 2017.

\bibitem{graves2006connectionist}
A.~Graves, S.~Fern{\'a}ndez, F.~Gomez, and J.~Schmidhuber.
\newblock Connectionist temporal classification: labelling unsegmented sequence
  data with recurrent neural networks.
\newblock In {\em Proceedings of the 23rd international conference on Machine
  learning}, pages 369--376. ACM, 2006.

\bibitem{grosicki2009icdar}
E.~Grosicki and H.~El~Abed.
\newblock Icdar 2009 handwriting recognition competition.
\newblock In {\em Document Analysis and Recognition, 2009. ICDAR'09. 10th
  International Conference on}, pages 1398--1402. IEEE, 2009.

\bibitem{he2016deep}
K.~He, X.~Zhang, S.~Ren, and J.~Sun.
\newblock Deep residual learning for image recognition.
\newblock In {\em Proceedings of the IEEE conference on computer vision and
  pattern recognition}, pages 770--778, 2016.

\bibitem{jaderberg2014synthetic}
M.~Jaderberg, K.~Simonyan, A.~Vedaldi, and A.~Zisserman.
\newblock Synthetic data and artificial neural networks for natural scene text
  recognition.
\newblock {\em arXiv preprint arXiv:1406.2227}, 2014.

\bibitem{karatzas2013icdar}
D.~Karatzas, F.~Shafait, S.~Uchida, M.~Iwamura, L.~G. i~Bigorda, S.~R. Mestre,
  J.~Mas, D.~F. Mota, J.~A. Almazan, and L.~P. de~las Heras.
\newblock Icdar 2013 robust reading competition.
\newblock In {\em Document Analysis and Recognition (ICDAR), 2013 12th
  International Conference on}, pages 1484--1493. IEEE, 2013.

\bibitem{kingma2014adam}
D.~Kingma and J.~Ba.
\newblock Adam: A method for stochastic optimization.
\newblock {\em arXiv preprint arXiv:1412.6980}, 2014.

\bibitem{liu2016star}
W.~Liu, C.~Chen, K.-Y.~K. Wong, Z.~Su, and J.~Han.
\newblock Star-net: A spatial attention residue network for scene text
  recognition.
\newblock In {\em BMVC}, 2016.

\bibitem{marti2002iam}
U.-V. Marti and H.~Bunke.
\newblock The iam-database: an english sentence database for offline
  handwriting recognition.
\newblock {\em International Journal on Document Analysis and Recognition},
  5(1):39--46, 2002.

\bibitem{poznanski2016cnn}
A.~Poznanski and L.~Wolf.
\newblock Cnn-n-gram for handwriting word recognition.
\newblock In {\em Proceedings of the IEEE Conference on Computer Vision and
  Pattern Recognition}, pages 2305--2314, 2016.

\bibitem{shi2017end}
B.~Shi, X.~Bai, and C.~Yao.
\newblock An end-to-end trainable neural network for image-based sequence
  recognition and its application to scene text recognition.
\newblock {\em IEEE transactions on pattern analysis and machine intelligence},
  39(11):2298--2304, 2017.

\bibitem{smith2007overview}
R.~Smith.
\newblock An overview of the tesseract ocr engine.
\newblock In {\em Document Analysis and Recognition, 2007. ICDAR 2007. Ninth
  International Conference on}, volume~2, pages 629--633. IEEE, 2007.

\bibitem{wang2011end}
K.~Wang, B.~Babenko, and S.~Belongie.
\newblock End-to-end scene text recognition.
\newblock In {\em Computer Vision (ICCV), 2011 IEEE International Conference
  on}, pages 1457--1464. IEEE, 2011.

\end{thebibliography}
}

\end{document}